\documentclass[letterpaper]{article}
\usepackage[submission]{aaai2026}  
\usepackage{times}  
\usepackage{helvet}  
\usepackage{courier}  
\usepackage[hyphens]{url}  
\usepackage{graphicx} 
\usepackage{natbib}  
\usepackage{caption} 
\usepackage{algorithm}
\usepackage{algorithmic}

\usepackage{newfloat}
\usepackage{listings}
\DeclareCaptionStyle{ruled}{labelfont=normalfont,labelsep=colon,strut=off} 
\floatstyle{ruled}
\newfloat{listing}{tb}{lst}{}
\floatname{listing}{Listing}
\usepackage{amsmath}
\usepackage{amssymb}
\usepackage{xcolor}     
\usepackage{pgfplots}
\usepackage{booktabs}
\usepackage{amssymb}
\usepackage{pifont}
\usepackage{hyperref}
\hypersetup{
  colorlinks=true,
  urlcolor=blue,        
  linkcolor=black,      
  citecolor=black
}

\newcommand{\cmark}{\ding{51}}  
\newcommand{\xmark}{\ding{55}}  

\pgfplotsset{compat=1.18}  
\newcommand{\modelname}{\textit{RadDiagSeg-M}\space}
\newcommand{\modelnamenoindent}{\textit{RadDiagSeg-M}}
\newcommand{\datasetname}{\textit{RadDiagSeg-D\space}}
\newcommand{\datasetnamenoindent}{\textit{RadDiagSeg-D}}

\title{\textit{RadDiagSeg-M}: A Vision Language Model for Joint Diagnosis and Multi-Target Segmentation in Radiology}
\author{
Chengrun Li\textsuperscript{\rm 1}\,
Corentin Royer\textsuperscript{\rm 3}\,
Haozhe Luo\textsuperscript{\rm 2}\,
    Bastian Wittmann\textsuperscript{\rm 1}\,
    Xia Li\textsuperscript{\rm 3} \\
    Ibrahim Hamamci\textsuperscript{\rm 1}\,
    Sezgin Er\textsuperscript{\rm 1}\,
    Anjany Sekuboyina\textsuperscript{\rm 1}\,
    Bjoern Menze\textsuperscript{\rm 1 \textdagger}
}
\affiliations{

    \textsuperscript{\rm 1} University of Zurich, Switzerland \\
\textsuperscript{\rm 2} University of Bern, Switzerland \\
\textsuperscript{\rm 3} ETH Zurich, Switzerland \\
\textsuperscript{\textdagger} Corresponding Author 
}

\usepackage{bibentry}

\begin{document}

\maketitle

\begin{abstract}
Most current medical vision language models struggle to jointly generate diagnostic text and pixel-level segmentation masks in response to complex visual questions. This represents a major limitation towards clinical application, as assistive systems that fail to provide both modalities simultaneously offer limited value to medical practitioners. To alleviate this limitation, we first introduce \datasetnamenoindent, a dataset combining abnormality detection, diagnosis, and multi-target segmentation into a unified and hierarchical task. \datasetname covers multiple imaging modalities and is precisely designed to support the development of models that produce descriptive text and corresponding segmentation masks in tandem.
Subsequently, we leverage the dataset to propose a novel vision-language model, \modelnamenoindent, capable of joint abnormality detection, diagnosis, and flexible segmentation. \modelname provides highly informative and clinically useful outputs, effectively addressing the need to enrich contextual information for assistive diagnosis. Finally, we benchmark \modelname and showcase its strong performance across all components involved in the task of multi-target text-and-mask generation, establishing a robust and competitive baseline. Code for this work is published at: \href{https://github.com/fitzlithepius/RadDiagSeg}{github.com/RadDiagSeg}
\end{abstract}

\section{Introduction}
Advances in medical Vision Language Models (VLMs), such as LLaVA-Med, Med-PaLM M, MedGemma, have demonstrated vast assistive potentials for several medical tasks~\cite{li2023llava, medpalm,  sellergren2025medgemma}. As one of the most important diagnostic tools, radiological images (\textit{e.g.}, X-ray, CT, and MRI) offer a high amount of clinical insights. Whilst demonstrating strong capabilities in understanding radiological images and answering questions, medical VLMs unanimously fail to accurately reflect their findings through an accurate pixel-level segmentation mask. This renders their results less reliable, given the known problem of LM hallucination~\cite{liu2024surveyhallucination}. To effectively assist clinicians, a model should be able to provide textual answers and accurate segmentation masks in tandem.

\begin{figure}[t!]
  \centering
   \includegraphics[width=\linewidth]{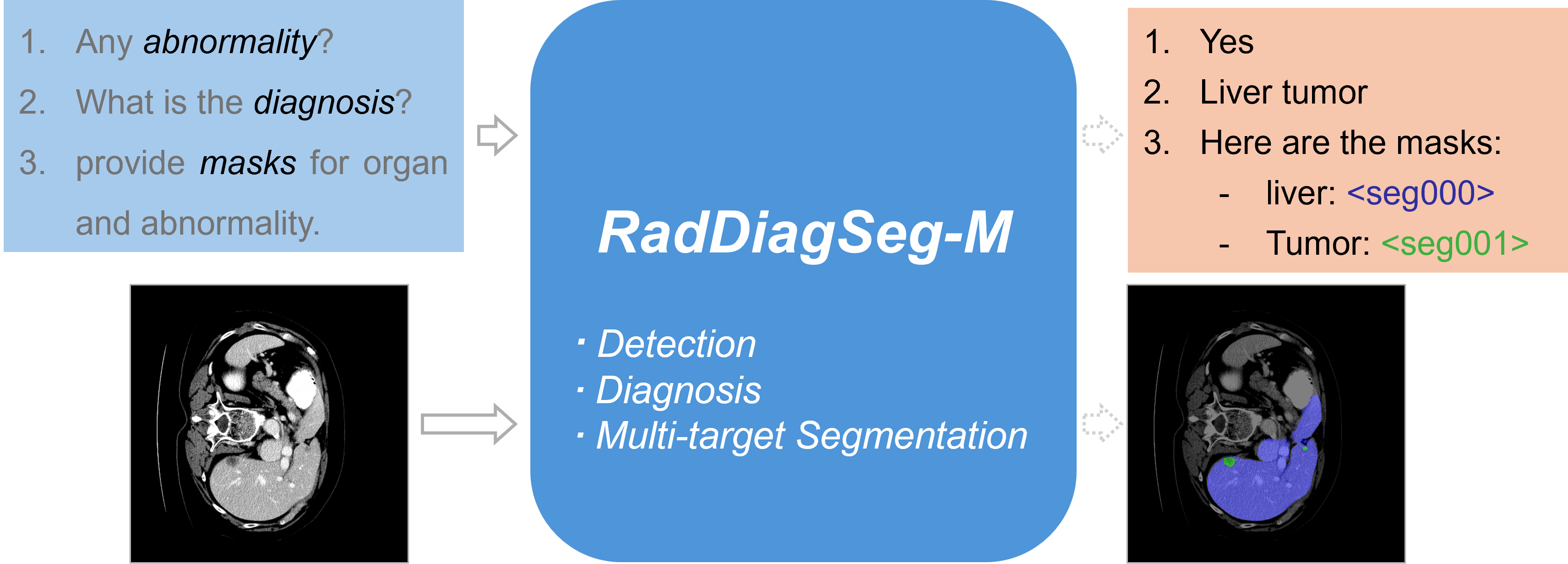}
   \caption{Overview. \modelname is capable of jointly detecting and diagnosing abnormality, and providing multi-target segmentation masks.}
   \label{fig:front-page}
\end{figure}

The emergence of promptable segmentation foundation models (FMs) in the medical field, such as BiomedParse~\cite{biomedparse} and MedSAM~\cite{medsam}, enables the segmentation of varying medical targets with user-defined prompts, \textit{e.g.}, points, boxes, and text labels. Architectures such as LISA~\cite{lisa} and Sa2VA~\cite{yuan2025sa2va} provide ways to connect powerful pre-trained VLMs with the segmentation FMs, enabling segmentation with free-form text prompts. Early endeavors in the medical field followed the idea of LISA, such as VividMed~\cite{luo2024vividmed} and MedPLIB~\cite{medplib}. However, these models only work with the Referring Segmentation (Ref-Seg) or the Visual Question Answering (VQA) task, thus failing at more complex tasks requiring both textual answers and masks at the same time. Furthermore, current models are unable to generate multiple masks for a given image with one prompt, partially compromising flexibility and clinical utility. Narrowing this gap, our model can answer complex questions with text and segmentation masks of abnormalities and the corresponding infected organs.

Given the complexity and novelty of our task, we identify the absence of datasets, an effective benchmark, and suitable models to serve as a baseline. In this paper, we address these limitations: First, we propose the \datasetname dataset consisting of more than 28k high-quality data samples covering major radiological modalities, \textit{i.e.}, X-ray and CT, by aggregating and processing several public datasets~\cite{tahir2021covidquex, biomedparse,antonelli2022medical}. Each sample in \datasetname consists of 3-step hierarchical questions: a close-ended VQA for abnormality detection, an open-ended VQA for diagnosis, and a segmentation task for one or multiple objects. The questions get more difficult with progression, and failing an earlier step will lead to automatic failure for the rest. The design of this task fosters explicit, step-by-step answers that are easier to inspect and offer more granularity, while maintaining extensive coverage of the VLM capabilities. Second, we propose the \modelname (\underline{Rad}iological \underline{Diag}nostic \underline{Seg}mentation) VLM. \modelname is built upon the state-of-the-art architecture proposed by ~\citet{lisa}, where we expand the vocabulary of our medical VLM with special segmentation generation tokens to trigger mask generation. Notably, the overall model is trained end-to-end with a unified training process. Unlike existing models, which only support single mask generation, our model inherently supports a flexible number of mask generations. Finally, we design the evaluation process and publish the benchmarking tool for the research community to enable effective and thorough evaluation for multi-target text-and-mask generation tasks.

\noindent
In summary, our contributions are as follows:
\begin{enumerate}
    \item We introduce \datasetnamenoindent, a dataset comprising over 28k samples. Each sample includes three-step hierarchical questions covering VQA and segmentation.
    \item We propose \modelnamenoindent, a radiological VLM that is capable of joint abnormality detection, diagnosis, and flexible multi-target segmentation. 
    \item We design and publish a benchmarking tool for the effective evaluation of \datasetnamenoindent.
    \item Experimental results indicate that \modelname achieves start-of-the-art results on the VQA sub-tasks of the \datasetname benchmark, while establishing a competitive baseline for the complete task.
\end{enumerate}

\section{Related Methods}
\label{sec:lr-review}
\subsection{Segmentation Models in Medical Images}
Specialist models targeting a specific organ of a specific modality have been the dominating approach regarding medical image segmentation over the past decade. CNN-based architectures like the U-Net~\cite{unet} and its variants, such as Swin-UNet, ResUNet++, and TransUnet~\cite{cao2022swin,jha2019resunet++, chen2021transunet} have achieved competitive results on many specific segmentation tasks. Whilst offering robust performance on the trained modality, such specialist models, however, have shown limited generalization across modalities. Recent universal segmentation paradigms~\cite{kirillov2023segment,zou2023segment} enabled the emergence of many generalist models for medical images, such as MedSAM~\cite{medsam}, SAM-Med2D~\cite{cheng2023sammed2d}, and BiomedParse~\cite{biomedparse}. Unlike the SAM-like models~\cite{kirillov2023segment,medsam,cheng2023sammed2d}, requiring explicit positional prompts such as dots or boxes, BiomedParse~\cite{biomedparse} works solely with textual labels. Despite its novelty, we argue that the text encoder employed by BiomedParse inherently lacks the ability to understand free-form text prompts. Therefore, it cannot handle the complex VQA tasks. \modelname overcomes this limitation with Multimodal LM to enable language comprehension and complex question-answering behavior.
\subsection{Medical Vision Language Models}
Vision Language Models (VLMs) in the general vision domains have demonstrated their promising abilities in the vision-related understanding and answering tasks, \textit{e.g.}, VQA and image captioning.~\cite {wang2024qwen2, gemma3, chen2024internvl}. As medical images play a vital role in clinical practices, medical VLMs such as RadFM~\cite{radfm}, LlaVA-Med~\cite{li2023llava}, MedPaLM M~\cite{medpalm}, Med-flamingo~\cite{moor2023med} were developed with large-scale radiology datasets and with techniques from the general domain, \textit{e.g.}, instruction fine-tuning~\cite{llava}. Despite strong performance on downstream tasks, existing models lack the ability to segment abnormalities—a critical requirement in radiological practice.

\subsection{Medical VLMs with Segmentation Mask Outputs}
The success of models such as LISA~\cite{lisa}, LLM-Seg~\cite{llmseg}, SegLLM~\cite{segllm}, and Sa2VA~\cite{yuan2025sa2va} demonstrates the potential of connecting VLMs with generalist segmentation models. In the medical domain, several VLMs~\cite{luo2024vividmed, medplib} inspired by LISA have been developed to perform segmentation tasks. However, these models cannot simultaneously answer complex questions and generate segmentation masks for relevant findings. Concurrent work of UniBiomed~\cite{wu2025unibiomed} attempts to address this challenge. Given the critical role of radiological imaging in medical diagnosis, a model that can detect, diagnose, and deliver pixel-level segmentations is essential for advancing toward a reliable radiological AI assistant. To this end, we propose \modelnamenoindent, a VLM enhanced with the capability to answer complex questions alongside accurate, pixel-level segmentation across major radiological imaging modalities.

\section{Methods}
\subsection{\modelnamenoindent: Model Architecture}
\label{sec:method-modelarc}
Our model generally follows the embedding-as-prompt architecture proposed in LISA~\cite{lisa}, which has been widely adopted by VLMs with segmentation capabilities in the medical field~\cite{luo2024vividmed, medplib}. However, most of the above-mentioned models utilize their components trained on in-house data or train a decoder module from scratch. Besides, training of these models involves multiple stages and (un-)freezing different parts at different stages. In our work, we propose a model structure built entirely with open-source components, trained in a simplistic yet elegant two-stage end-to-end process. Our model surpasses LISA-like models regarding flexibility in mask generation and the joint language-segmentation capability, as is demonstrated in Table~\ref{tb:model-capabilities}. 

\begin{figure}[t!]
  \centering
   \includegraphics[width=\linewidth]{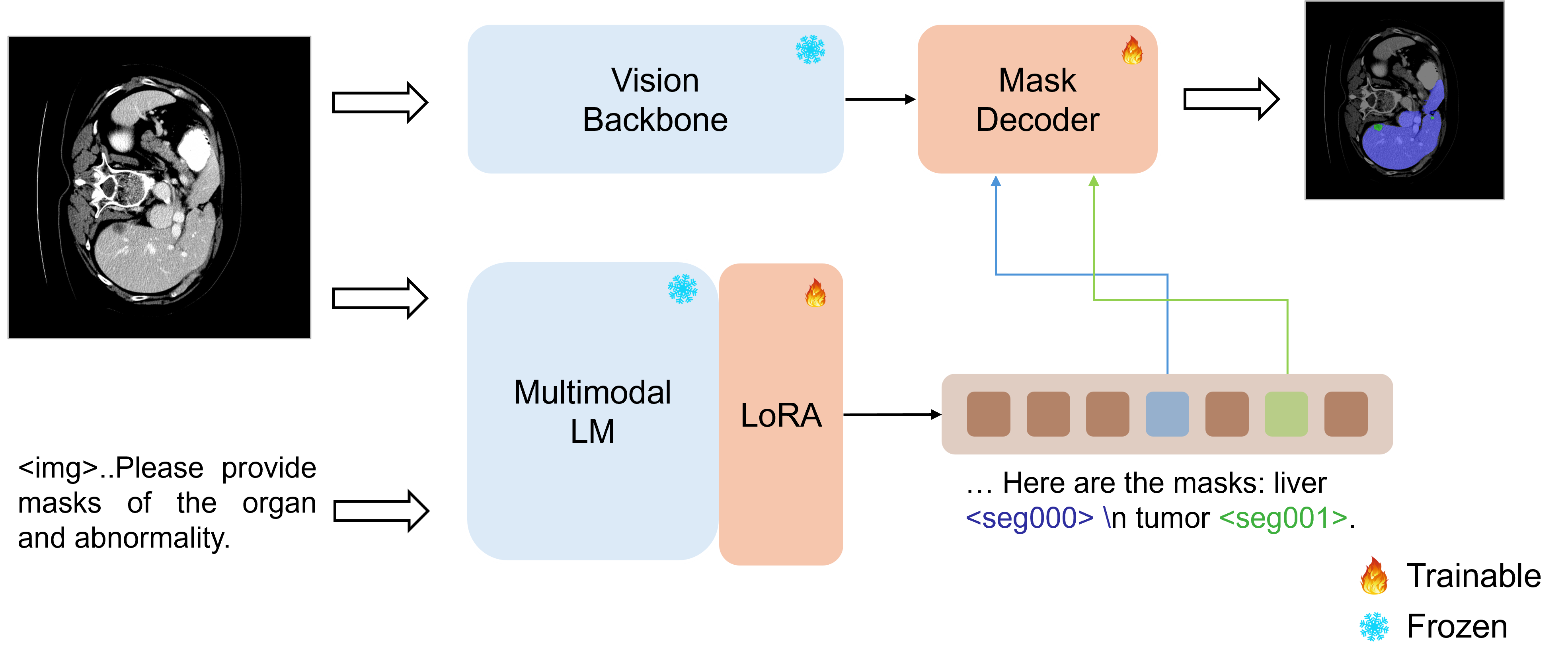}
   \caption{Architecture of \modelnamenoindent. Our model supports multi-target flexible segmentation.}
   \label{fig:model-struc}
\end{figure}

\modelname consists of three main components: a vision backbone, a multimodal language model (multimodal LM), and a mask decoder, as is shown in Figure~\ref{fig:model-struc}. The multimodal LM processes a user text prompt together with an image to generate a text answer. Following LISA, we re-purpose a series of \textit{\textless seg\textgreater} tokens to guide the segmentation process. If the multimodal LM chooses to carry out the segmentation task, a special segmentation token, \textit{e.g.}, \textit{\textless seg000\textgreater}, is generated. The last layer hidden embedding of the special token is passed through the mask decoder to create a binary segmentation mask. These special tokens also differentiate the normal VQA behavior from tasks requiring segmentation output, since only the answer containing \textit{\textless seg\textgreater} will activate the segmentation process.

\paragraph{Vision backbone.} The vision backbone $F_{\text{enc}}$ extracts pixel-level visual features from the input medical image $x_{\text{image}}$ to support mask generation. We adopt the image encoder from MedSAM~\cite{medsam} to leverage pretrained model knowledge. Given a batch of $b$ input images \( x_{\text{image}} \in \mathbb{R}^{b \times 3 \times W \times H} \), the images are transformed by the vision backbone into image embeddings  \( z_{\text{image}} \in \mathbb{R}^{b \times 256 \times \frac{W}{16} \times \frac{H}{16}} \).

\paragraph{Multimodal LM.} Many general domain multimodal LMs demonstrate strong question answering capabilities when directly applied to medical tasks (see LLaVA~\cite{llava}, Qwen-VL~\cite{wang2024qwen2}, and InternVL~\cite{chen2024internvl}). However, due to the unique properties of radiological images, Multimodal LM's internal image encoders trained on natural images typically fail to generalize. Therefore, to create a powerful multimodal LM for radiological images, we substitute the multimodal LM's native image encoder with a pretrained medical CLIP-based variant~\cite{zhang2024biomedclip}. Additionally, we apply LoRA~\cite{hu2022lora} for parameter-efficient fine-tuning of LM. We discuss design choices in the ablation studies. 

Every sample in a batch of $b$ consists of an image prompt and a text prompt: $(x_{\text{image}}, x_{\text{text}})$. We feed the image-text pair to the multimodal LM, which in turn outputs a text response $\hat{y}_\text{text}$. The corresponding last-layer hidden state can be described as $z_\text{emb}$, the process of which can be formulated as
\begin{equation}
  z_\text{emb} = \text{LM}(x_{\text{image}}, x_{\text{text}}).
  \label{eq:mlm}
\end{equation}

\paragraph{Mask decoder.} The mask decoder module consists of the mask decoder from MedSAM~\cite{medsam} and a linear projection layer to align the embedding shape. When the multimodal LM decides to generate segmentation mask(s), $\hat{y}_\text{text}$ will contain one or multiple segmentation control tokens. Therefore the last-layer embedding $z_\text{emb}$ contains a non-empty subset of segmentation token embeddings $h_\text{seg}$, expressed as \( z_{\text{emb}} \supseteq \{ h_{\text{seg}}^{(i)} \}_{i=1}^{k},0 < k \leq n \).

\begin{table}[!t]
\centering
\setlength{\tabcolsep}{3pt}
\begin{tabular}{l|ccccc}
\toprule
 & \textbf{Dect} & \textbf{Diag} & \textbf{Seg}  & \textbf{Mul-Seg}& \textbf{VQA-Seg} \\
\midrule
BiomedParse     & \xmark &\xmark                &\cmark     &\cmark &\xmark \\
LISA     & \cmark  & \xmark  &\cmark     &\xmark & \xmark  \\
MedGemma    & \cmark & \cmark &\xmark &\xmark&\xmark \\
MedPLIB      & \cmark & \cmark               &\cmark  &\xmark   &\xmark  \\
UniBiomed    & \cmark & \cmark               &\cmark  &\xmark   &\cmark   \\
\midrule
\modelname   & \cmark & \cmark  & \cmark & \cmark & \cmark\\
\bottomrule
\end{tabular}%
\caption{Comparison of model capabilities on detection, diagnosis, segmentation, multi-target segmentation, and VQA segmentation. Mul-Seg refers to the ability to generate multiple clearly-referred masks for different targets. \modelname is the first model capable of all tasks.}
\label{tb:model-capabilities}
\end{table}

Iteratively for each of the segmentation token embedding $h_\text{seg}^{(i)}$, the mask decoder MD will process the image embedding $z_\text{image}$ to generate the binary segmentation mask $\hat{M}$. The decoding process for a mask can be formalized as:
\begin{equation}
\hat{M} = \text{MD}(h_\text{seg}^{(i)}, z_\text{image}).
\label{eq:mask_decoder}
\end{equation}

\paragraph{Training objectives.} The training objective is to jointly minimize the auto-regressive LM loss $\mathcal{L}_\text{text}$ and the segmentation loss $\mathcal{L}_\text{seg}$. As the segmentation loss, we adopt a combination of pixel-level binary cross-entropy (BCE) loss and Dice loss, 
following MedSAM~\cite{ma2021loss}. Overall, the composition objective $\mathcal{L}$ can be framed as:

\begin{equation}
\mathcal{L} = \lambda_{\text{text}} \cdot \mathcal{L}_{\text{text}} + \lambda_{\text{seg}} \cdot \mathcal{L}_{\text{seg}},
\label{eq:total-loss}
\end{equation}

\noindent where
\begin{equation}
\mathcal{L}_{\text{seg}} = \lambda_{\text{bce}} \cdot \mathcal{L}_{\text{bce}}(\hat{M}, M) + \lambda_{\text{dice}} \cdot \mathcal{L}_{\text{dice}}(\hat{M}, M).
\label{eq:seg-loss}
\end{equation}


\subsection{\modelnamenoindent: Training Data and Tasks}
\label{sec:model-data-compo}
As is illustrated in Figure~\ref{fig:dataset}, the training process of our model involves three different tasks, all of which are derived from widely-adopted public datasets. 
In the following, we describe these three tasks in detail:

\begin{figure*}[t!]
  \centering
\includegraphics[width=0.8\linewidth]{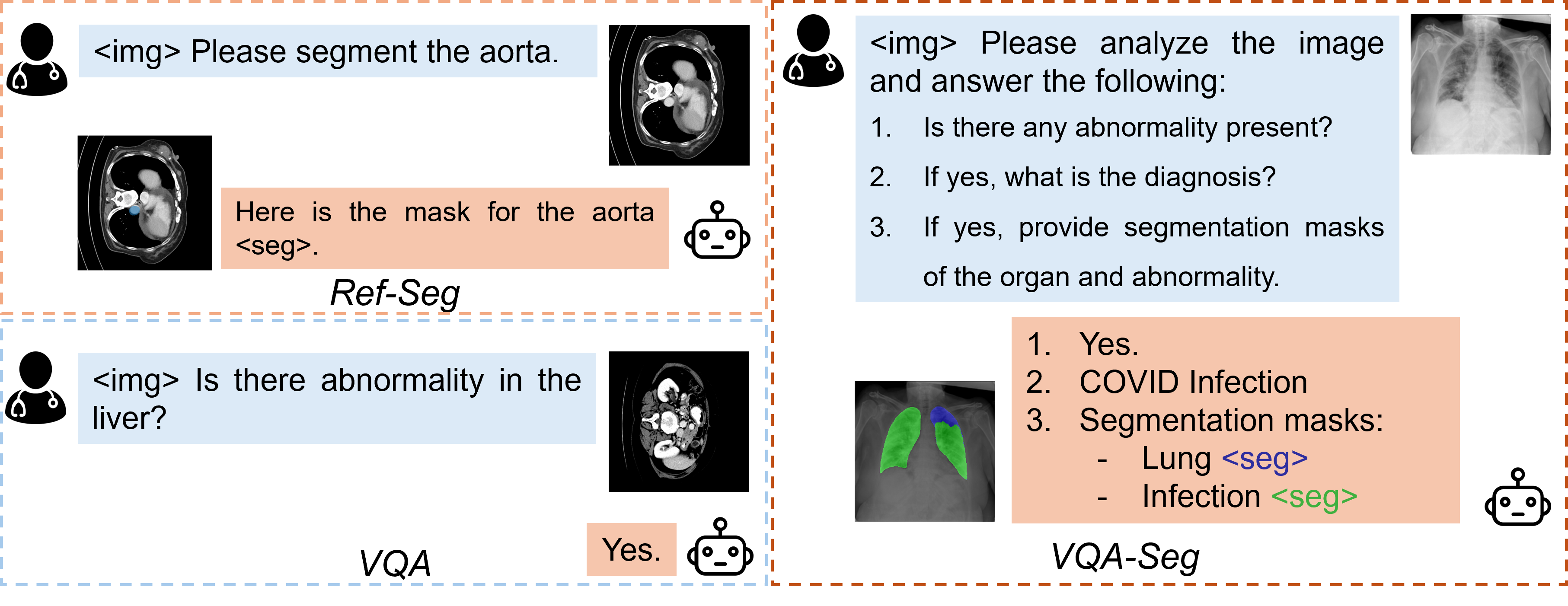}
   \caption{Three types of tasks in training. Notably, the VQA Segmentation (VQA-Seg) is a novel complex task that involves three steps: textual tasks of detection and diagnosis, and a multi-target segmentation task.}
   \label{fig:dataset}
\end{figure*}

\paragraph{Referring segmentation task (Ref-Seg).} A sample contains an image, a binary segmentation mask, and a text label for the segmentation target. A template to normalize our data points is: \textbf{USER}: ``\textless img\textgreater \space Please segment \textit{Target} in the \textit{Image Modality}." \textbf{ASSISTANT}: ``Here is the mask for \textit{Target} \textless seg\textgreater." The text label can either directly refer to the segmentation target, \textit{e.g.}, liver, or by its functionality, \textit{e.g.}, hepatic organ. During training, a label will be randomly sampled to ensure diversity in training data. We adopt subsets of BiomedParseData~\cite{biomedparse} as the training set.

\paragraph{Visual question answering task (VQA).} VQA task involves the generation of accurate natural language answers to visually-based questions. To preserve the visual-language capability of pretrained multimodal LM according to~\citet{mckinzie2024mm1}, we incorporate public radiological VQA datasets throughout the training. We utilize VQA-RAD~\cite{vqarad} and SLAKE~\cite{liu2021slake}, both of which provide high-quality visually-based question answer pairs in the radiology domain.

\paragraph{VQA segmentation task (VQA-Seg).} To foster \modelnamenoindent's ability in answering complex questions while providing segmentation mask(s), we proposed a complex task comprised of hierarchical VQA tasks and segmentation tasks. Data is structured in a unified format, where a data sample with positive findings is formatted as shown in Figure~\ref{fig:dataset}. For data samples with negative findings (no visible abnormality), a negative text answer is utilized. Every question includes three steps: a close-ended question with a binary answer for detection, an open-ended question for diagnosis, and a subsequent segmentation task requiring one or multiple masks. Specifically, we process a subset from BiomedParseData~\cite{biomedparse, antonelli2022medical} and COVID-QU-Ex~\cite{tahir2021covidquex} as the \datasetname dataset used in experiments. 
Notably, our proposed VQA-Seg task guides the model to think and respond in a step-by-step manner, producing both coherent diagnoses and clinically meaningful segmentation masks. 

\subsection{\datasetname Benchmark}
Our proposed VQA-Seg task is complex and challenging in its structured steps of questions. For the text part, we consider the right combination of detection and diagnosis as a correct prediction. Since the mask generation is conditioned on text embedding, we argue that a correct text answer is a prerequisite for meaningful segmentation masks. 

The hierarchical complexity of the task also presents challenges to the effective evaluation, since there hasn't been an established benchmark for this composite task. We address this deficiency by extending a widely adopted medical benchmarking tool. The evaluation process treats the first step of the task as a close-ended question and computes the F1 score. For the second step, we treat the diagnosis problem as an open-ended VQA and report the overall F1, \textit{i.e.}, only the correct combination of detection and diagnosis is considered a success. A failure in the earlier stage will automatically stop the evaluation, leading to zero results in the following stages. For example, if the model answers, \textit{e.g.}, \textit{``1. Yes.  2. There is ..."} while the ground truth is \textit{``1. No"}. The answer fails at the detection level, leading automatically to the failure of the following steps and the whole task. We document details of the evaluation process in the appendix. 

\label{sec:exp}
\section{Experiments}
\subsection{Model Implementation}
\label{sec:exp-setting}
\paragraph{Model details.} Unless otherwise specified, the implementation of \modelname relies on the following components: We adopted the respective MedSAM components as the vision backbone and mask decoder. For the multimodal LM, there are three key components: the LM, the image encoder, and the multimodal projector. We use the LM from \textit{PaliGemma2-3b-pt-224}~\cite{paligemma}. The image encoder is BiomedCLIP~\cite{zhang2024biomedclip} due to its existing knowledge on medical images and CLIP-style pretraining~\cite{radford2021learning}. A multimodal projector projects the raw image embedding to the LM embedding~\cite{tolstikhin2021mlp}. In comparison to previous works, all of our components are open-sourced, underpinning the effectiveness and flexibility of our design. The choices of components are discussed in the ablation studies. Additional details can be found in the appendix.

\begin{table*}[t]
\centering
\begin{tabular}{@{}l c c c c c c c c@{}}
\toprule
 & \multicolumn{4}{c}{\textbf{VQA‑RAD}} & \multicolumn{4}{c}{\textbf{SLAKE}} \\
\cmidrule(lr){2-5} \cmidrule(lr){6-9}
 & F1 & Recall & OpenQ‑Acc & OpenQ‑Recall
 & F1 & Recall & OpenQ‑Acc & OpenQ‑Recall \\
\midrule
RadFM             & \underline{0.442} & \textbf{0.474} & \underline{0.335} & \underline{0.407}
                  & 0.714             & 0.695           & \underline{0.725} & \underline{0.758} \\
LlaVA‑Med         & 0.069             & 0.372           & 0.140              & 0.246
                  & 0.075             & 0.443           & 0.362            & 0.492            \\
MedGemma          & 0.164             & 0.449           & \textbf{0.415}     & \textbf{0.518}
                  & 0.066             & 0.565           & 0.593            & 0.664            \\
LISA              & 0.052             & 0.229           & 0.080              & 0.132
                  & 0.070             & 0.314           & 0.207            & 0.244            \\
UniBiomed         & 0.020             & 0.084           & 0.045              & 0.023
                  & 0.106             & 0.011           & 0.076            & 0.088            \\

PaliGemma2        & 0.352             & 0.350           & 0.150              & 0.169
                  & 0.337             & 0.336           & 0.245            & 0.253            \\\midrule
\modelnamenoindent‑PT   & \textbf{0.460}    & \underline{0.458} & 0.238            & 0.291
                  & \underline{0.718} & \underline{0.716} & 0.666          & 0.705            \\
\modelnamenoindent‑FT   & 0.351             & 0.353           & 0.306              & 0.245
                  & \textbf{0.774}    & \textbf{0.778}   & \textbf{0.754}  & \textbf{0.779}   \\
\bottomrule
\end{tabular}
\caption{VQA results for \modelnamenoindent. “OpenQ” denotes open‑ended questions, while “Acc” denotes accuracy. \modelname outperforms most baselines by a wide margin and achieves state‑of‑the‑art performance on SLAKE.}
\label{tb:vqa}
\end{table*}

\paragraph{Training details.} We utilize two NVIDIA H100 (80 GB) GPUs for training and leverage the DeepSpeed ZeRO engine~\cite{rasley2020deepspeed} for efficient distributed computation. We adopt a two-stage training process, which comprises initial pre-training followed by fine-tuning. The pre-training stage aims at aligning different components and activating the full model capability for segmentation, whereas the fine-tuning stage optimizes model performance on the specific VQA-Seg task. As indicated in Figure~\ref{fig:model-struc}, the trainable components are the LoRA, the multimodal projector within the multimodal LM, the mask decoder, together with the text embeddings of segmentation tokens. Complete specification is documented in the appendix. 

\paragraph{Datasets.} Given the efforts required in the annotation of radiological images and the scarcity of medical data in general, there are few datasets providing both pixel-level masks for the organ and abnormality. We transformed and aggregated COVID-QU-Ex~\cite{tahir2021covidquex} and subsets of MSD~\cite{antonelli2022medical, biomedparse} into \datasetnamenoindent, containing 22k samples for training and 6.8k for testing. For X-ray, potential abnormalities are COVID-19 and non-COVID infection, where healthy samples represent negatives. For CT, potential abnormalities are liver and pancreas tumors, where the slices with no visible tumor are considered as negative samples. We format the samples according to the example shown in Figure~\ref{fig:dataset}. 

 The training process involves three types of tasks, where a different combination is adopted for the two stages. For the pre-training stage, we adopt Ref-Seg task with 199k samples from BiomedParseData~\cite{biomedparse} covering all three key modalities of radiology images, \textit{i.e.}, X-ray, CT and MRI. 5k VQA samples are taken from VQA-RAD~\cite{vqarad} and SLAKE~\cite{liu2021slake} to maintain the joint understanding capability. In the fine-tuning stage, we adopt a mixture of the VQA and the \datasetname datasets, resulting in a total of 32k high-quality samples. \datasetname constitutes 22k samples, and the rest 10k are VQA samples. We discuss the effect of the data mix of the fine-tuning stage in the ablation studies. For 3D modalities like CT and MRI, the slices from the same volume don't infiltrate across data partitions~\cite{biomedparse}.

\paragraph{Evaluation metrics.} For evaluation, we adopt F1 and Recall as metrics for the VQA tasks. We additionally document the Recall and Accuracy for open-ended questions. Following common practices, the Dice score is used to benchmark the quality of segmentation. For \datasetnamenoindent, given the label imbalance, we use F1 as the metric for detection and diagnosis. 

\subsection{Experiment Results}
\modelname is novel in its ability to answer complex questions with structured text answers and well-referred segmentation mask(s). We first present the model's capability on downstream tasks of Ref-Seg and VQA. Subsequently, we present the results on the complex task of VQA-Seg.

\begin{figure*}[!t]
  \centering
\includegraphics[width=0.80\linewidth]{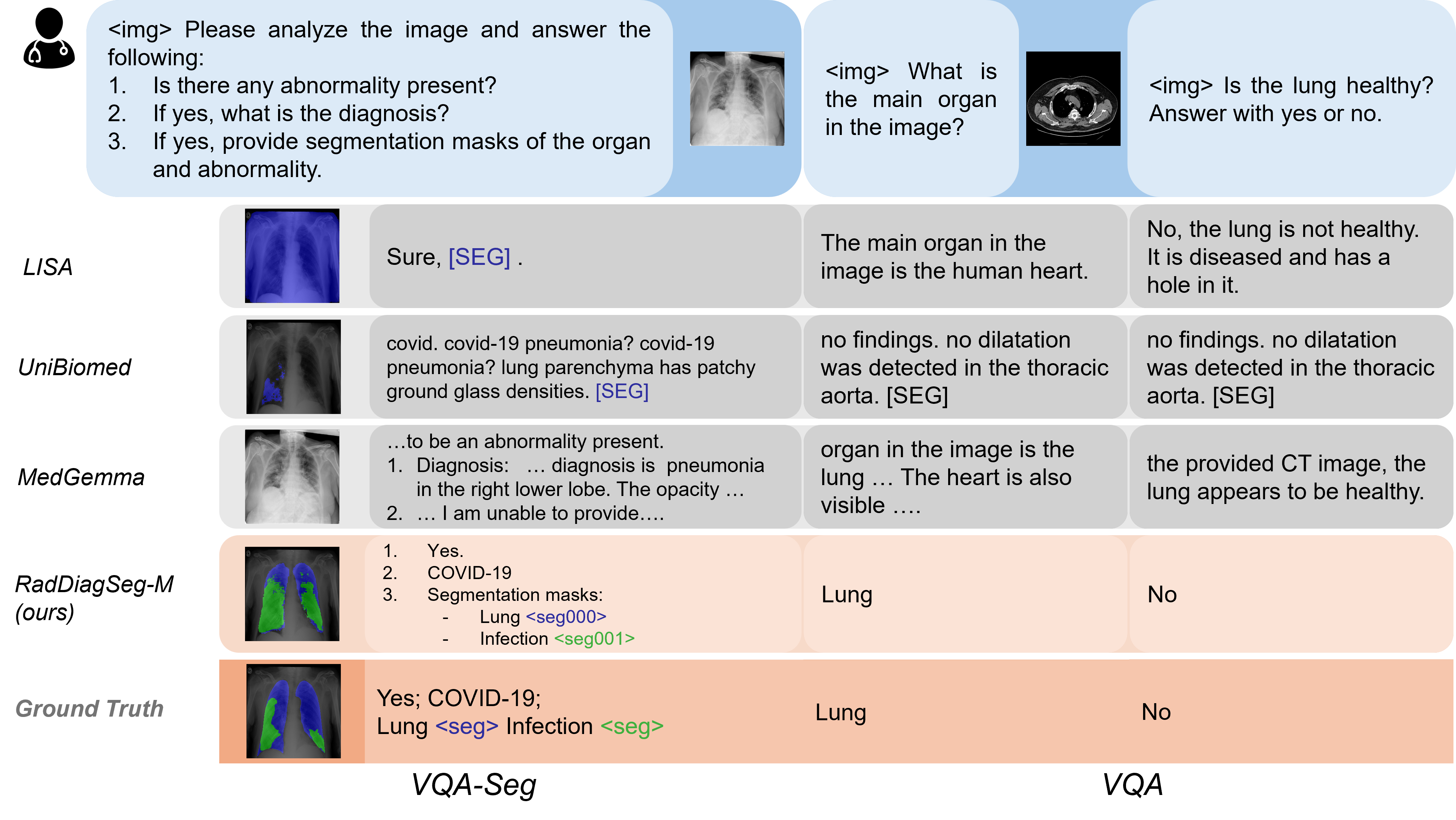}
  \caption{Qualitative results for the VQA-Seg and VQA tasks. \modelname is the only model answering all three questions correctly while providing multiple well-referred masks following the instructions. The results underpin the capability of \modelname at multi-target text-and-mask generation.}
  \label{fig:qualitative-analysis}
\end{figure*}

\begin{figure}[ht]
  \centering
\includegraphics[width=1\linewidth]{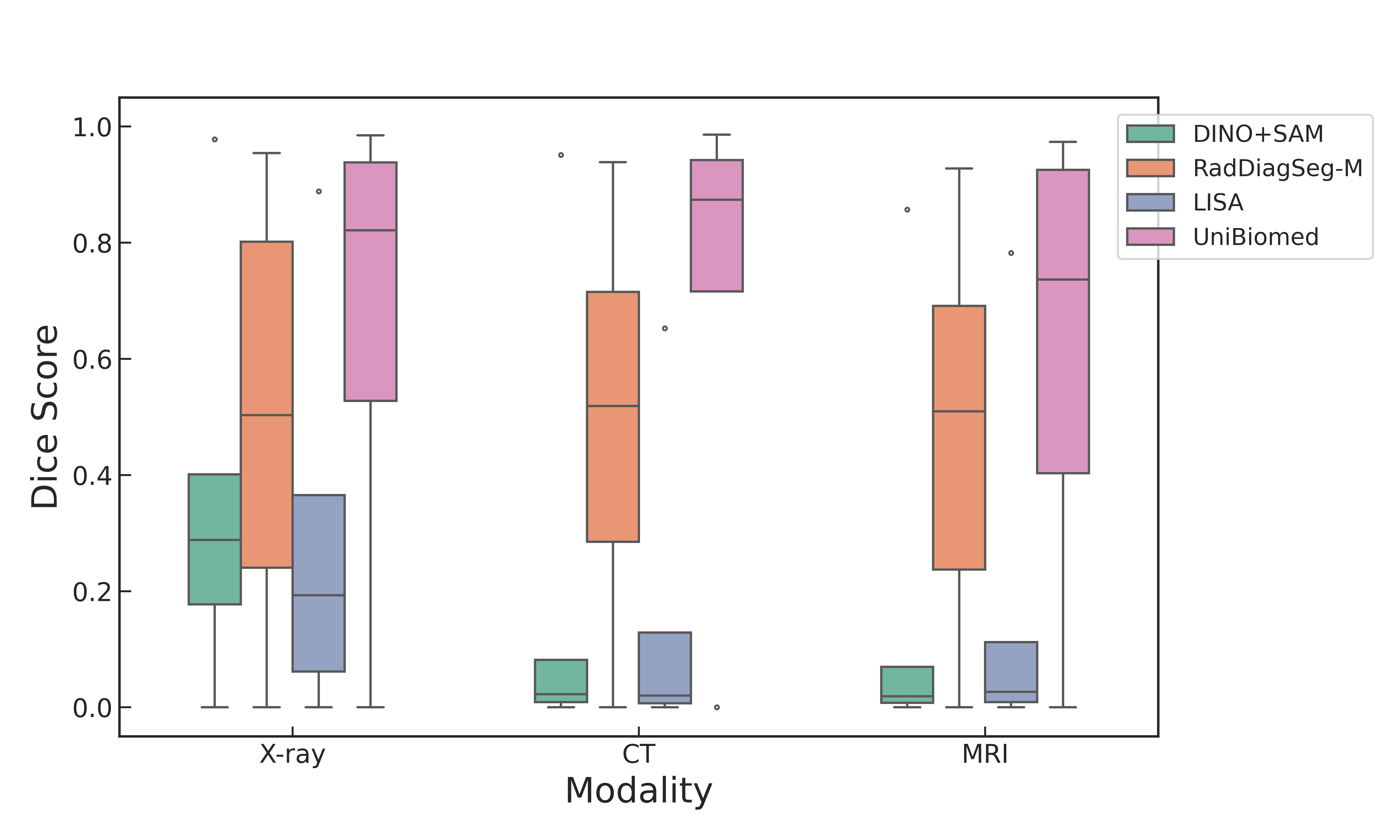}
   \caption{Box plot comparing performances of models on the Ref-Seg task across three modalities. \modelname significantly outperforms the baseline methods.}
   \label{fig:refseg}
\end{figure}

\paragraph{Ref-Seg results.} To evaluate the effectiveness of the pre-training stage, we benchmarked on the Ref-Seg task across all three modalities, \textit{i.e.}, X-ray, CT, and MRI. As shown in Figure~\ref{fig:refseg}, we compare our pre-trained (PT) model with three approaches: the vanilla approach combining Grounding DINO with MedSAM~\cite{groundingdino, medsam}, LISA~\cite{lisa}, and UniBiomed~\cite{wu2025unibiomed}. In the first approach, Grounding-DINO processes a text label and generates a bounding box to prompt MedSAM for mask generation. LISA achieves strong performance across many general-domain benchmarks. We consider these two approaches as our baselines.

\modelname consistently outperforms both baselines across all modalities with a margin of over 0.2 in Dice. Our improvement confirms the advantages of the joint embedding space and the benefits of domain-specific pre-training. We acknowledge the gap between our model and the concurrent method UniBiomed. Despite the difference in training scale, we aim to create a model capable of joint text and multi-target segmentation generation (see Figure~\ref{fig:qualitative-analysis}). 

\paragraph{VQA results.} To maintain and improve the capabilities of image understanding and question answering, we have included proportions of VQA data in every training stage. Table~\ref{tb:vqa} presents the evaluation results on the test sets of two radiology VQA datasets: VQA-RAD~\cite{vqarad} and SLAKE~\cite{liu2021slake}. We benchmarked two variants of our model (pre-trained variant (PT), fine-tuned variant (FT)) against state-of-the-art (SOTA) medical VLMs (RadFM, LLaVA-Med, and MedGemma) and VLMs with segmentation capability (LISA, UniBiomed). Both variants of our model show improved performance over the base model PaliGemma2~\cite{paligemma}. We highlight that our FT model achieves state-of-the-art results on SLAKE. We observe a decline in part of the metrics from the PT to the FT variant, which we attribute to the joint learning objective of the \datasetname task. 

Notably, we observe the collective poor performance of current SOTA models with segmentation capabilities, LISA, and UniBiomed. Their scores indicate failures to correctly answer the majority of questions. Especially, UniBiomed was trained on both datasets, yet it demonstrates the worst performance reported. We present qualitative examples of close- and open-ended questions in Figure~\ref{fig:qualitative-analysis}. We analyze and discuss the results further in the qualitative analysis.

\paragraph{VQA-Seg results.} Given the novelty and complexity of the task, Table~\ref{tb:model-capabilities} indicates that \modelname is the only model capable of this task. To guarantee a fair comparison, if another tested model failed to generate meaningful results for the full task, we amputated the task to retrieve meaningful scores. Despite the efforts, empty fields in Table~\ref{tb:vqa-seg} indicate the failure of existing models to perform all the required sub-tasks. \modelname is the first model capable of joint detection, diagnosis, and multi-target segmentation.

Notably, comparable VLMs with segmentation capabilities demonstrate difficulties in following instructions and answering questions step-by-step. For example, even adding an explicit prompt requiring a binary \textit{yes/no} answer in the detection task, UniBiomed still fails to generate a binary answer. Similarly, LISA fails at generating meaningful text after the detection step. Towards the objective of building an assistive model for clinicians, such models' failure hinders the potential of clinical application. The result also verifies the necessity of more complex tasks like \datasetnamenoindent.

Besides being the first model capable of performing the whole task, we establish a robust baseline for all the tasks. On X-ray, \modelname shows leading performance in all the task categories. On CT, \modelname achieves comparable performance on detection, and an improved performance in diagnosis and organ segmentation. Collectively, we outperform existing methods with improved metrics and set a competent baseline for the task.

\begin{table*}[t!]
\centering
\begin{tabular}{@{}l cc cc cc cc@{}}
\toprule
& \multicolumn{4}{c}{\textbf{X-ray}} & \multicolumn{4}{c}{\textbf{CT}} \\
\cmidrule(lr){2-5} \cmidrule(lr){6-9}
& Detection & Diagnosis & \multicolumn{2}{c}{Segmentation}
& Detection & Diagnosis & \multicolumn{2}{c}{Segmentation} \\
& F1 & F1 & Dice-Org & Dice-Abn
& F1 & F1 & Dice-Org & Dice-Abn \\
\midrule
MedGemma          & \underline{0.904} & 0.625 &    – &    – & \textbf{0.526} & 0.336 &    – &    – \\
LISA\textsuperscript{*}
                  & 0.857             &    –   &    – &    – & \underline{0.521} &    –   &    – &    – \\
UniBiomed\textsuperscript{*}
                  & 0.838             & \underline{0.724} &    – & \underline{0.410}
                                     & 0.502 & \underline{0.627} &    – & \textbf{0.214} \\
\modelname        & \textbf{0.912}    & \textbf{0.864}
                  & \textbf{0.833} & \textbf{0.541}
                  & 0.506 & \textbf{0.657}
                  & \textbf{0.670} & \underline{0.103} \\
\bottomrule
\end{tabular}

\vspace{2mm}
\begin{minipage}{0.95\textwidth}
\footnotesize
\textsuperscript{*}Amputation and adaptation of questions needed for meaningful value.\\
– Model is \textit{not} capable of performing the task.
\end{minipage}
\caption{Performance on \datasetname task. “Org” abbreviates organ, and “Abn” abnormality. \modelname demonstrates state-of-the-art results on X-ray, and achieves comparably competitive results on CT, setting a robust baseline for the task.}
\label{tb:vqa-seg}
\end{table*}

\begin{table}[t]
\centering
\begin{tabular}{@{}l c c| c c@{}}
\toprule
\textbf{Encoder} & \textbf{LM} & \textbf{ImgTok} & \textbf{SLAKE} & \textbf{Ref‑Seg CT} \\
                 & params        & \#              & F1             & Dice            \\
\midrule
SigLIP           & 3b          & 256             & 0.740          & 0.227           \\
B‑CLIP           & 3b          & 256             & 0.744          & \textbf{0.488}  \\
MedSAM           & 3b          & 256             & 0.693          & 0.263           \\
\midrule       
B‑CLIP           & 10b         & 1024            & \textbf{0.763} & \underline{0.484}           \\
B‑CLIP           & 10b         & 256             & \underline{0.759}          & 0.483           \\
\bottomrule
\end{tabular}
\caption{Ablation of components in the multimodal LM, including image encoder, language model parameter count, and number of image tokens. Notably, the image encoder has the strongest overall impact on the downstream tasks. 
}
\label{tb:ablation-mllm}
\end{table}

\paragraph{Qualitative analysis.} Figure~\ref{fig:qualitative-analysis} illustrates the joint complex question answering and flexible segmentation abilities of \modelnamenoindent. For the VQA-Seg task, other comparable models, except MedGemma, fail to follow the instructions to answer the questions. More importantly, for LISA and UniBiomed, the failure in diagnosis subsequently leads to ambiguity in the segmentation target. As reported in UniBiomed~\cite{wu2025unibiomed}, its improvement of performance is partly attributed to the mask generation process conditioned on the textual output and input. Therefore, if the textual answer is ambiguous, the mask will also be less credible. Similarly, the complete devoid of useful textual information from LISA leads to ambiguity of the segmentation target. Furthermore, the example of the VQA task demonstrates the deterioration of language capability from UniBiomed, answering different questions with the same irrelevant and incorrect answer. Collectively, these examples confirm our claim that a more complex task with both text and segmentation, \textit{e.g.}, \datasetnamenoindent, is needed towards building an assistive VLM for radiological diagnosis.

Through the qualitative analysis, we demonstrate the practical significance of a complex VQA task like \datasetname for the assistive diagnosis in the radiological image field. We confirm through the success of \modelname that the joint ability to answer complex questions and generate multiple masks is both important and learnable.

\subsection{Ablation Studies}
\label{sec:ablation}

\paragraph{Effect of component choice in multimodal LM.} Starting with the image encoder in Table~\ref{tb:ablation-mllm}, using a medical-aware vision encoder B-CLIP (BiomedCLIP) significantly improves segmentation performance and moderately enhances performance on the visual-language understanding tasks, compared to the baseline of SigLIP. This highlights the importance of a vision encoder pre-trained with domain-specific images. While MedSAM offers a structurally simpler alternative, its lack of language awareness during pretraining appears to limit performance, particularly in cross-modal tasks.

\begin{table}[htbp]
\centering
\begin{tabular}{@{}cc| cc@{}}
\toprule
\textbf{Seg} & \textbf{Text} & \textbf{SLAKE} & \textbf{\datasetnamenoindent} \\
     \%             &        \%            & F1             & Diagnosis F1       \\
\midrule
0.8               & 0.2                & 0.694          & 0.812              \\
0.6               & 0.4                & \textbf{0.743} & \textbf{0.885}     \\
\bottomrule
\end{tabular}
\caption{Ablation of fine‑tuning data composition. “Seg” denotes the proportion of VQA‑Seg samples, and “Text” that of VQA samples. A higher proportion of “Text” samples improves the model's performance on downstream tasks.}
\label{tb:ablation-data}
\end{table}

Increasing the language model size and the number of image tokens further boosts performance following the scaling law~\cite{kaplan2020scaling}. Comparing variants with BiomedCLIP as encoder, we find that increasing the LM size and the image tokens yields consistent improvements on VQA tasks. However, scaling up doesn't result in any improvement on the Ref-Seg task.

\paragraph{Effect of data composition in fine-tuning stage.}
The objective of the fine-tuning stage is to jointly improve the performance on text generation and segmentation capabilities for the \datasetname task. We ran the ablation study on the X-ray portion of \datasetname and VQA datasets. Given the composite loss function of Equation~\ref{eq:total-loss} used in our training process, the task composition has a direct impact on the flow of gradients, thus directly influencing the outcomes. We ablated the effect in Table~\ref{tb:ablation-data}, where a higher proportion of VQA data with pure-text output mitigates the model collapsing on the language abilities, thus benefiting the joint improvement of all downstream tasks with a 0.07 increase on the \datasetname Diagnosis F1 score.

\section{Conclusion}
In this paper, we focus on the capabilities of radiological VLMs to jointly generate high-quality diagnostic text and clearly referred segmentation masks. To this end, we introduce \datasetnamenoindent, a dataset spanning major radiology modalities. The complex task of \datasetname is designed to improve the joint question answering and flexible segmentation abilities of medical VLMs. We further present \modelnamenoindent, a radiological VLM capable of joint abnormality detection, diagnosis, and flexible multi-target segmentation. Given the novelty of \datasetnamenoindent, we additionally release a benchmarking tool to support standardized evaluation. Experiments demonstrate that our model achieves competitive performance on downstream tasks and SOTA performance on SLAKE. Furthermore, \modelname is the first model capable of tackling the full complex task of \datasetnamenoindent, setting benchmarks on text-based tasks and establishing a strong baseline for the whole task.
\bibliography{aaai2026}

\newpage
\appendix
\section{Appendix}
\section{Dataset Details}
We provide more details on the three tasks we adopted in the training stages: Ref-Seg, VQA, and VQA-Seg.
\subsection{Ref-Seg Dataset. }Consisting of 199,829 samples, Ref-Seg includes diverse samples from three different radiological modalities. Table~\ref{tb:refseg-overview} presents the composition of the dataset.

\begin{table*}[htbp]
\centering
\begin{tabular}{l|ccp{5cm}}
\toprule
      & \textbf{Dataset Name}            & \textbf{Num Samples} & \textbf{Segmentation Target} \\ \midrule
X-ray & Chest Xray Masks and Labels Dataset &      1,632             &  healthy lung, tuberculosis lung                    \\ 
      & SIIM-ACR Pneumothorax Segmentation &         2379          &     pneumothorax                 \\ 
      & COVID-19 Radiography Database &       39,014            &    COVID, lung, viral pneumonia, lung opacity                  \\ \midrule
CT    & MSD                     &         27,699          & liver, liver tumour, pancreas, pancreas tumour, lung tumour, colon cancer primaries, spleen     \\ 
      & amos22                  &           108,704        & abdominal organs: spleen, right kidney, left kidney, gallbladder, esophagus, liver, stomach, aorta, inferior vena cava, pancreas, right adrenal gland, left adrenal gland, duodenum, bladder, prostate/uterus \\ 
      & COVID-19 CT            &1,187& lungs, covid-19 infections      \\ 
      & LIDR-IDRI               &           7,389        & lung nodule\\ \midrule
MRI   & amos22                  &          11,825         & abdominal organs as above     \\
\bottomrule
\end{tabular}
\caption{Overview of Ref-Seg Dataset. The majority of data points originate from the CT modality, followed by X-ray and MRI.}
\label{tb:refseg-overview}
\end{table*}

\subsection{VQA Dataset} VQA-RAD~\cite{vqarad} consists of 1,790 samples in the training set and 451 samples in the test set. We adopt the English subset of SLAKE~\cite{liu2021slake}, which comprises 4,919 samples in the training set and 1,061 samples in the test set.

\subsection{VQA-Seg Dataset} 
\begin{figure*}[htbp]
  \centering
  \includegraphics[width=0.75\linewidth]{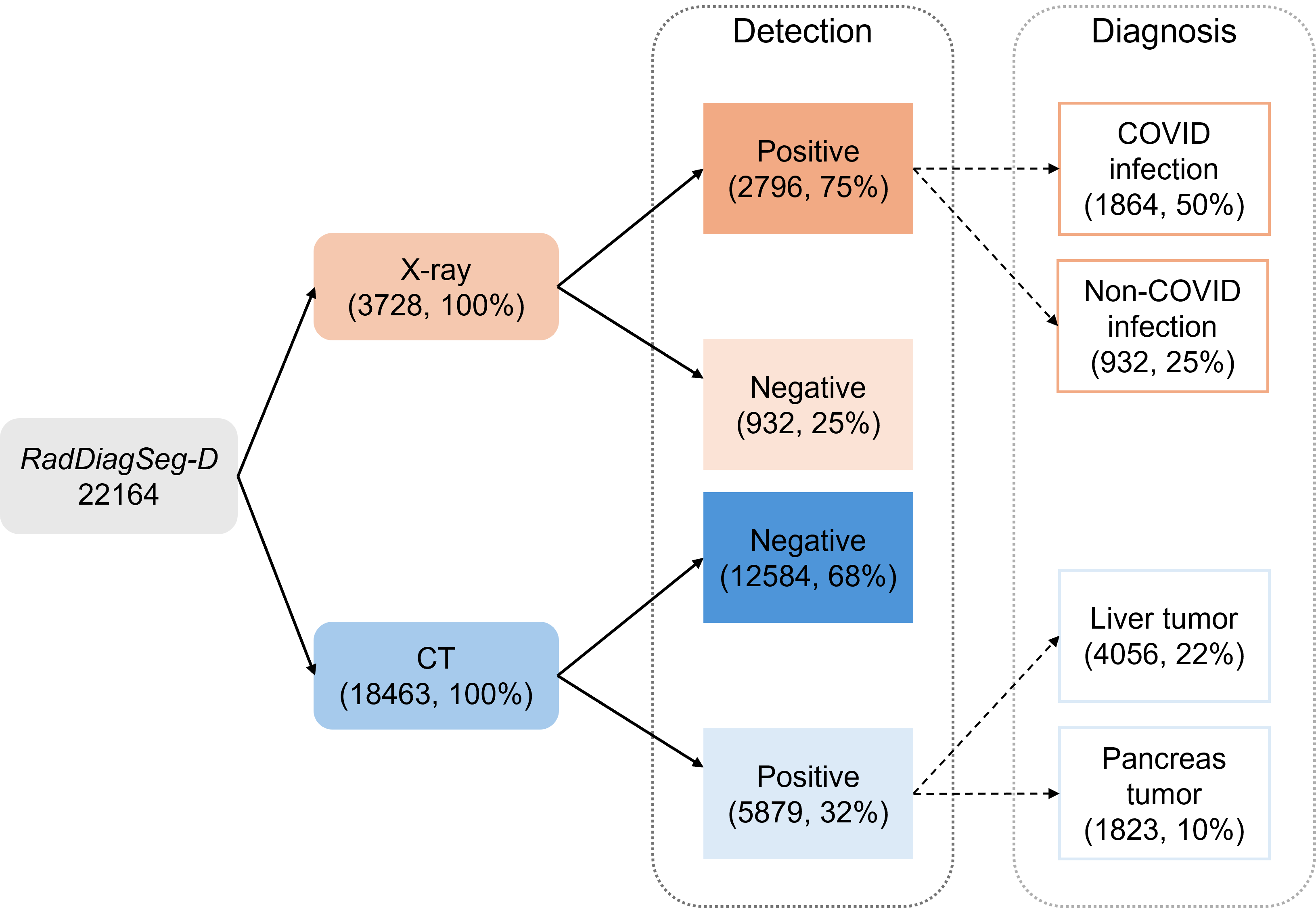}
  \caption{\datasetname overview and label distribution. The label imbalance within the training set poses another challenge to the task.}
  \label{fig:radvqa-seg}
\end{figure*}

We processed two subsets of MSD~\cite{antonelli2022medical, biomedparse} and COVID-QU-EX~\cite{tahir2021covidquex} into the \datasetname dataset. The resulting dataset contains over 28,000 samples, with 22k used for training and approximately 6k for testing. Figure~\ref{fig:radvqa-seg} illustrates the detailed composition of the \datasetname training set.

Label imbalance is observed in both imaging modalities. In the X-ray subset, positive labels are more prevalent than negative ones, whereas in the CT subset, the opposite trend is present. This imbalance increases the difficulty of the task, as models cannot rely on overfitting to a dominant class to achieve strong performance. Accordingly, we account for this factor in the evaluation and report the F1 score to enable a fair comparison across methods.

\section{Evaluation of \datasetname}
\datasetname consists of three-step questions: a close-ended VQA detection question, an open-ended VQA diagnosis question, and a multi-target segmentation task. During the evaluation, we evaluate the answers following the steps.

\subsection{Detection F1} Detection task evaluates on a binary basis, with positive (\textit{yes}) and negative (\textit{no}) findings. For the predicted results, we explicitly map the answer to the binary labels. The computation of the F1 score can be formalized as follows:
Given a set of binary true labels $y_{\text{true}} \in \{0, 1\}^n$ and processed predicted labels $y_{\text{pred}} \in \{0, 1\}^n$, we compute the F1 score:
\[
\text{F1} = 2 \cdot \frac{\text{Precision} \cdot \text{Recall}}{\text{Precision} + \text{Recall}}
\]
where the binary F1 score is the harmonic mean of precision and recall:
\[
\text{Precision} = \frac{TP}{TP + FP} \text{ \space }
\text{Recall} = \frac{TP}{TP + FN}
\]

\subsection{Diagnosis F1}
The second open-ended question, which pertains to diagnosis, presents a greater challenge. Our evaluation focuses on the correctness of the diagnostic answer, hence the overall correct text answer to the whole task. We consider a diagnosis correct if it meets one of the following two situations:
\begin{itemize}
    \item Ground truth is negative; processed detection prediction is also negative.
    \item Ground truth is positive; processed detection prediction is positive, \textit{and} diagnostic label is present in the predicted text.
\end{itemize}
Otherwise, an answer is considered wrong. For each instance $i$, we formalize the process as a binary correctness indicator:
\[
c^{(i)} = \begin{cases}
1 & \text{if } y_\text{pred} = y_\text{true} \\
0 & \text{otherwise}
\end{cases}
\]
The F1 score is then computed between the predicted correctness $\{c^{(i)}\}$ and a ground truth vector of ones, reflecting whether the model answers both the detection and diagnosis questions correctly.

\subsection{Segmentation Dice} The last step of the task is the evaluation of segmentation masks. We report the mean Dice score of true positive valid predictions in the paper (see Table~\ref{tb:vqa-seg}). The Dice score aims to evaluate the overlap between predicted segmentation and ground truth, with a range from 0 (no overlap) to 1 (perfect overlap). Mathematically, for a binary-class segmentation, the Dice score is defined as:

\[
\text{Dice} = 2 \cdot \frac{ |P \cap G|}{|P| + |G|}
\]

where:
\begin{itemize}
    \item \( P \) is the set of predicted foreground pixels,
    \item \( G \) is the set of ground truth foreground pixels,
    \item \( |P \cap G| \) is the number of correctly predicted foreground pixels (i.e., true positives).
\end{itemize}

\subsection{Amputated Evaluation}
Amputated evaluation is employed when no meaningful results can be observed by directly applying the standardized evaluation procedure as introduced above. In such cases, we decompose the question into three evaluation steps: \textit{Detection}, \textit{Diagnosis}, and \textit{Diagnosis + Segmentation}. For numbers reported in Table~\ref{tb:vqa-seg}, LISA failed after the \textit{Detection} task. UniBiomed struggled to follow instructions to answer \textit{yes/no} in the \textit{Detection} step and generated diagnostic text, thus requiring extra steps for meaningful data analysis. 

\section{Implementation Details}
\subsection{Details of Multimodal LM}
 
\begin{figure*}[htbp]
  \centering
  \includegraphics[width=0.4\linewidth]{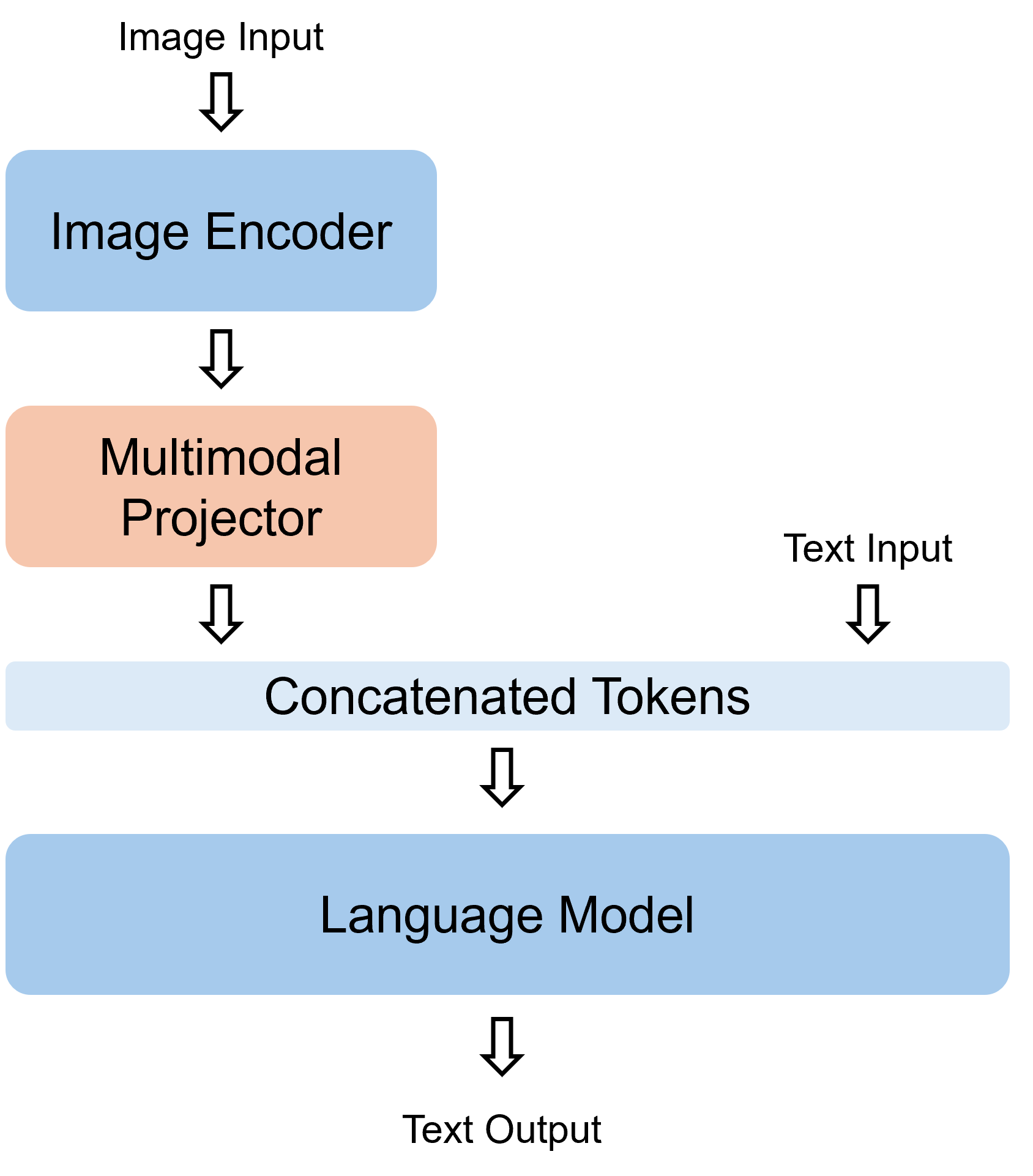}
  \caption{Structure of Multimodal LM. Multimodal projector aligns the visual and textual embedding space and reduces the number of image tokens. }
  \label{fig:mllm}
\end{figure*}

Figure~\ref{fig:mllm} illustrates the architecture of the Multimodal Language Model (Multimodal LM) and the flow of multimodal information. As the core component responsible for integrating diverse input modalities and making crucial decisions, a robust Multimodal LM serves as the foundation of \modelnamenoindent. While previous sections have detailed the image encoder and language model, we now provide a more focused discussion on the multimodal projector, which bridges vision and language modalities.

The multimodal projector performs two key functions: aligning the embedding dimensions and reducing the number of image tokens. Given an image input $x_\text{image}$, its encoded representation is denoted as $z'_\text{image} \in \mathbb{R}^{N_{\text{image}} \times D_{\text{image}}}$, where $N$ is the number of tokens and $D$ the embedding dimension. Similarly, the text input $x_\text{text}$ yields an embedding $z_\text{text} \in \mathbb{R}^{N_{\text{text}} \times D_{\text{text}}}$ after the language model's embedding layer.

We denote the projection function as $f_\text{proj}$ and the target number of image tokens as $\hat{N}_\text{image}$. The transformation of the multimodal projector is given by:
\begin{equation}
    \hat{z}'_{\text{image}} = f_{\text{proj}}(z'_{\text{image}}) \in \mathbb{R}^{\hat{N}_{\text{image}} \times D_{\text{text}}}
\end{equation}

In our implementation, we draw inspiration from the architecture of MLP-Mixer~\cite{tolstikhin2021mlp}, applying projection along both the token and embedding dimensions, with an intermediate transposition step. Specifically:
\begin{itemize}
    \item \textbf{Embedding dimension alignment ($D$)}: Following prior works~\cite{llava, mckinzie2024mm1}, we employ a two-layer multilayer perceptron (MLP) to map visual features into the language embedding space.
    \item \textbf{Token reduction ($N$)}: To condense image information, we adopt a Perceiver-style cross-attention module~\cite{jaegle2022perceiver}, using a fixed set of learnable queries to extract a compressed representation.
\end{itemize}

\begin{table*}[htbp]
\centering
\begin{tabular}{lcc}
\toprule
              & \multicolumn{1}{c}{Pre-training} & \multicolumn{1}{c}{Fine-tuning} \\ \cmidrule{2-3} 
Scheduler     & \multicolumn{2}{c}{Warmup + Cosine}                             \\
Optimizer     & \multicolumn{2}{c}{AdamW\cite{adam}}                                     \\
Loss          & \multicolumn{2}{c}{$\lambda_{\text{text}}=1.0$ \space $\lambda_{\text{seg}}= 1.0$}                                        \\
& \multicolumn{2}{c}{$\lambda_{\text{dice}}=0.5$ \space $\lambda_{\text{bce}}= 2.0$}\\
Num Trainable Params          & \multicolumn{2}{c}{625M}                                        \\
epochs        & 3                             & 5                            \\
learning rate & 2e-4                          & 2e-5                          \\
batch size    & 256                           & 64                           \\
training  time (hrs) & 70  & 20 \\
\bottomrule
\end{tabular}
\caption{Training specification in the pre-training and fine-tuning stages.}
\label{tb:train-params}
\end{table*}

\subsection{Training Parameters}
Table~\ref{tb:train-params} summarizes the complete training configuration employed in our experiments. For both the pre-training and fine-tuning stages, we validate on the test set every 100 steps and select the best-performing checkpoint based on validation performance as the final model. Specifically, the \textit{Pre-training (PT)} variant corresponds to the final checkpoint at the end of training, while the \textit{Fine-tuning (FT)} variant is taken from the checkpoint at step 1000.

\section{Discussion on Ambiguity of Model Answers}
Figure~\ref{fig:qualitative-analysis} presents the outputs of comparable models, highlighting the deteriorated language capabilities of UniBiomed. As noted in the UniBiomed paper~\citep{wu2025unibiomed}, its performance improvements are partly attributed to the generation of segmentation masks conditioned on both textual output and user input. However, when the textual response is ambiguous or misleading, the reliability of the generated mask correspondingly degrades. In such cases, as exemplified in Figure~\ref{fig:qualitative-analysis}, potential users may find it difficult to interpret the reference of the predicted mask, reducing its clinical utility. Similarly, the complete lack of informative text in LISA's output results in ambiguity regarding the segmentation target. Collectively, these examples substantiate our claim that complex multimodal tasks, such as those presented in \datasetnamenoindent, are essential for developing truly assistive VLMs for radiological diagnosis.

From the perspective of clinical assistance, UniBiomed’s response in Figure~\ref{fig:qualitative-analysis} demonstrates that even if the segmentation mask appears accurate, it holds limited value for clinicians if the textual reference is unclear. In contrast, \modelname not only provides explicit labels for the predicted mask but also includes contextual information, such as an additional organ mask, offering greater potential for clinical support and interpretability.

\section{More Qualitative Examples}
\begin{figure*}[htbp]
  \centering
  \includegraphics[width=0.8\linewidth]{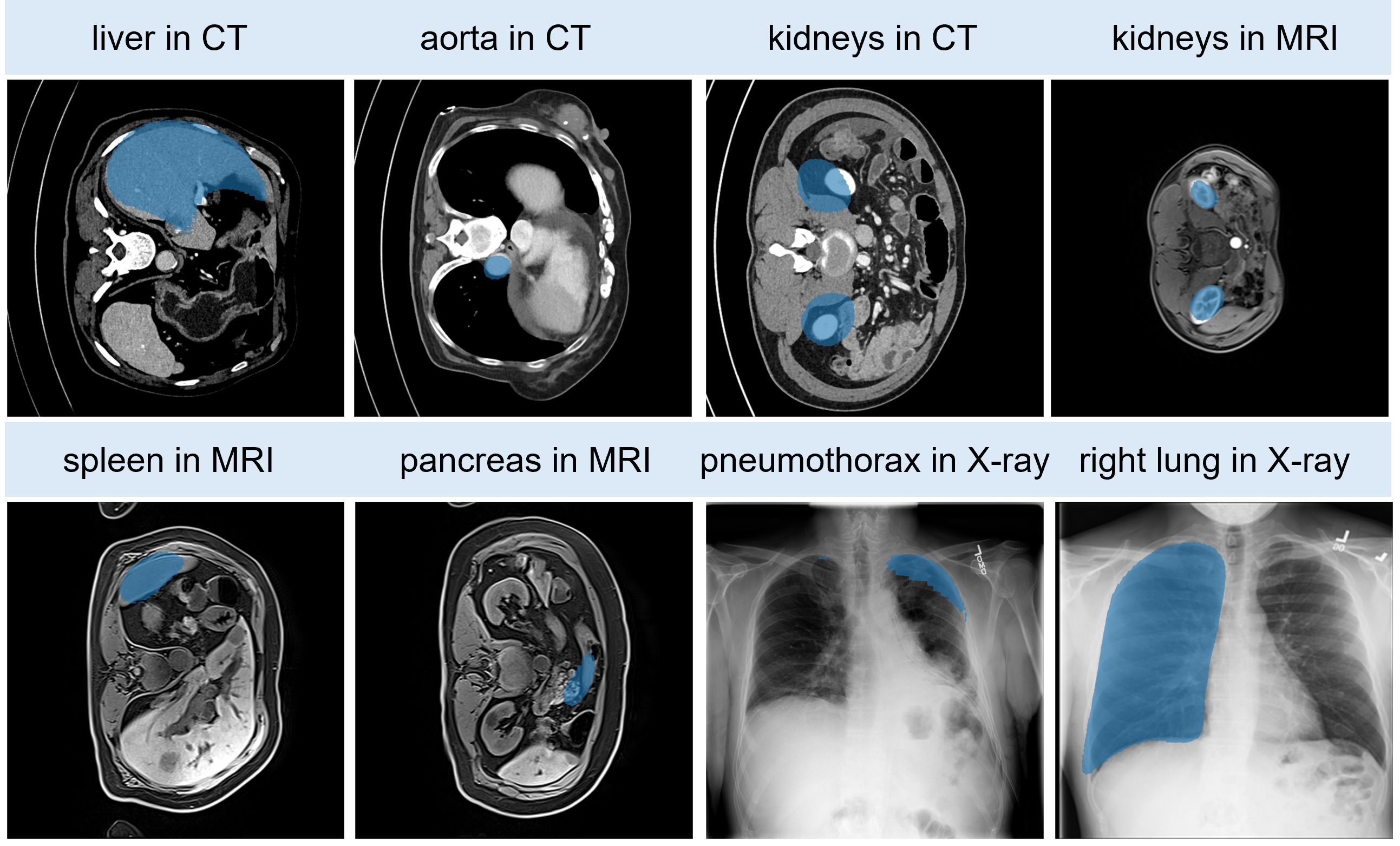}
  \caption{Qualitative Examples of Ref-Seg. \modelname provides accurate segmentation masks across radiological modalities: X-ray, CT, and MRI. However, the model struggles to accurately segment targets with irregular shapes, e.g. pancreas.}
  \label{fig:ref-seg-examples}
\end{figure*}

\begin{figure*}[htbp]
  \centering
  \includegraphics[width=0.9\linewidth]{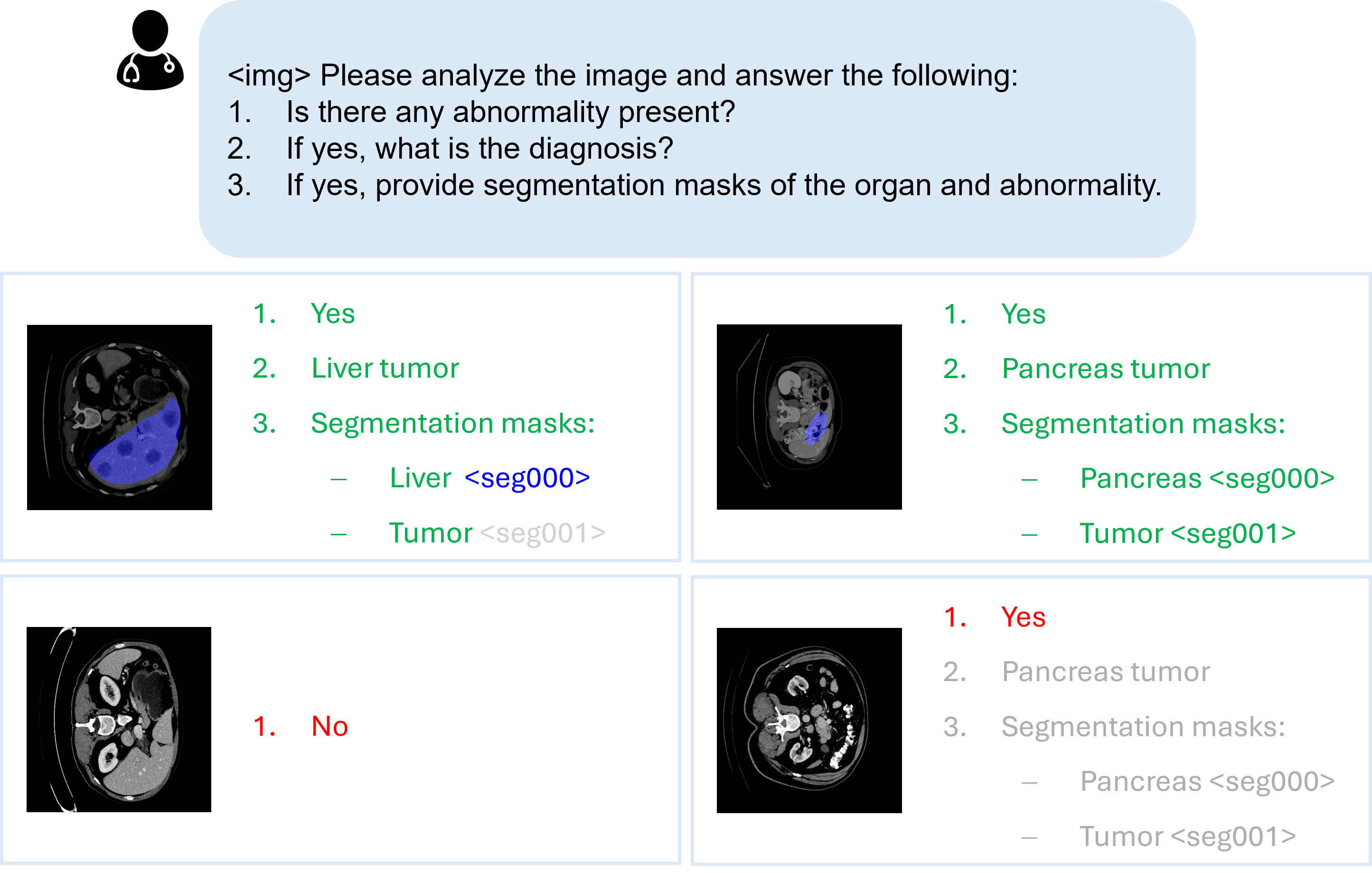}
  \caption{Qualitative Examples of \datasetname in CT. Answers are from \modelname with the organ mask visualized. \textit{green} marks correct textual answer, while \textit{red} the wrong answer. Notably, if the model fails at the detection step, as in the right-bottom example, evaluation will automatically end.}
  \label{fig:ct-vqaseg-examples}
\end{figure*}

\begin{figure*}[htbp]
  \centering
  \includegraphics[width=0.9\linewidth]{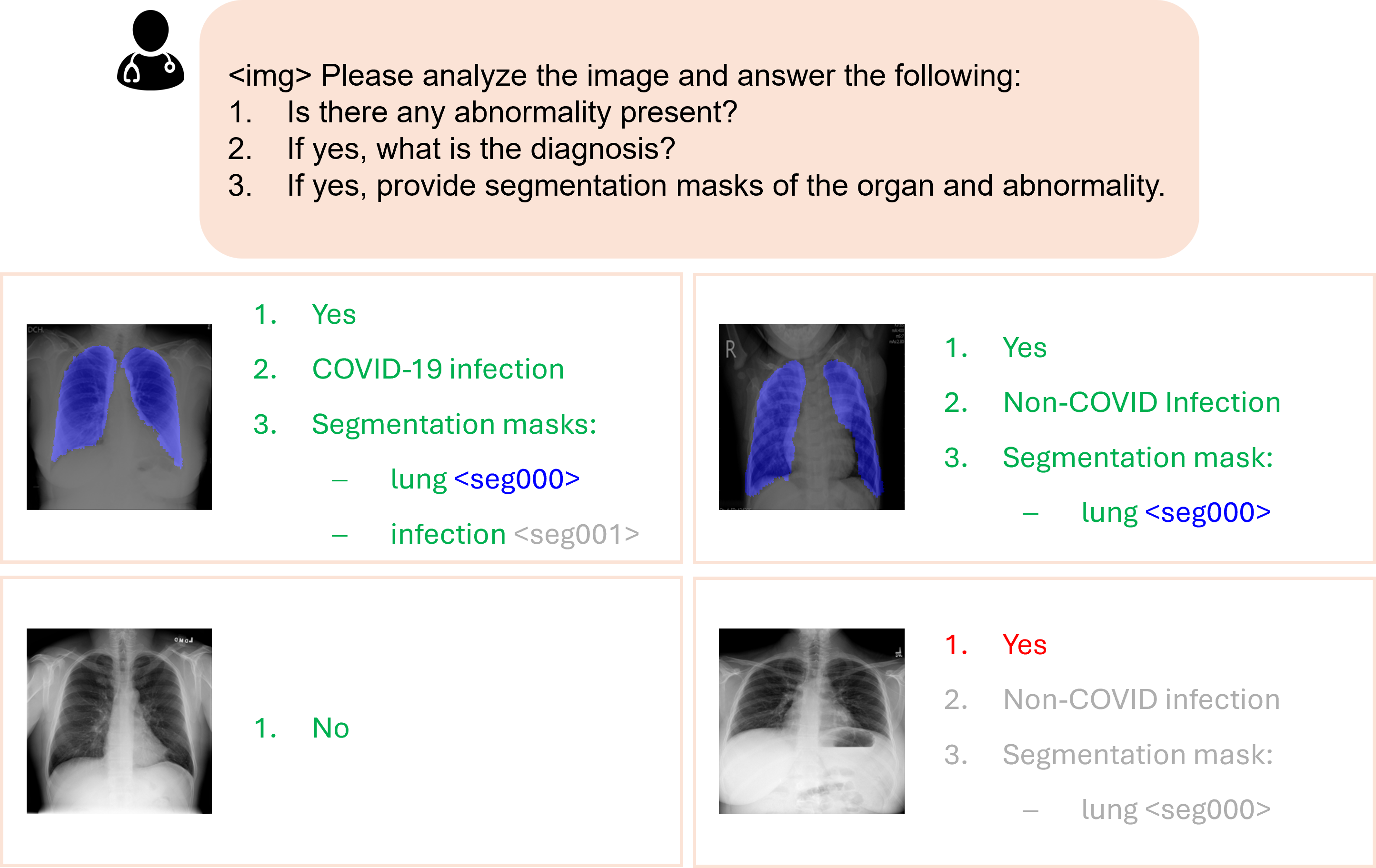}
  \caption{Qualitative Examples of \datasetname in X-ray. Answers are from \modelname with the organ mask visualized. \textit{green} marks correct textual answer, while \textit{red} the wrong answer.}
  \label{fig:xray-vqaseg-examples}
\end{figure*}

We present more qualitative examples of Ref-Seg (Figure~\ref{fig:ref-seg-examples}) and \datasetname (Figure~\ref{fig:ct-vqaseg-examples} for CT and Figure~\ref{fig:xray-vqaseg-examples} for X-ray). We observe that \modelname generally provides accurate and context-aware segmentation masks in the Ref-Seg task. Examples from \datasetname demonstrate further the \modelnamenoindent's capability in answering complex questions and providing multi-target segmentation. We also present examples where our model fails in both detection and diagnosis. These failure cases highlight the current limitations of the model and offer insights for future improvement.

\section{Limitations}
While we believe the introduction of \datasetname and \modelname represents a significant step toward the development of assistive radiological  VLMs that provide meaningful clinical support, we acknowledge certain limitations. In particular, \datasetname is subject to label variability, primarily due to the limited availability of open-source datasets that provide both diagnostic text and multi-target segmentation masks. Additionally, there remains room for improvement in segmentation performance, especially for small or subtle anatomical targets. Addressing these challenges—particularly enhancing joint complex question-answering and fine-grained segmentation—constitutes a key direction for our future work.

\end{document}